\documentclass[11pt]{article}

\usepackage[letterpaper, top=1.05in, bottom=1.05in,
            left=1.1in, right=1.1in]{geometry}
\usepackage{amsmath, amssymb}
\usepackage{graphicx}
\usepackage{booktabs}
\usepackage{array}
\usepackage{xcolor}
\usepackage[hidelinks, colorlinks=true,
            linkcolor=blue!60!black,
            citecolor=blue!60!black,
            urlcolor=blue!60!black]{hyperref}
\usepackage{microtype}
\usepackage{enumitem}
\usepackage{setspace}
\usepackage{fancyhdr}
\usepackage{titlesec}
\usepackage{caption}
\usepackage{placeins}

\titleformat{\section}{\large\bfseries}{\thesection.}{0.5em}{}[\vspace{-0.3em}\rule{\linewidth}{0.4pt}\vspace{0.2em}]
\titleformat{\subsection}{\normalsize\bfseries}{\thesubsection.}{0.5em}{}
\titlespacing{\section}{0pt}{1.4ex plus 0.4ex minus 0.2ex}{0.6ex plus 0.2ex}
\titlespacing{\subsection}{0pt}{1.0ex plus 0.3ex}{0.4ex plus 0.1ex}
\definecolor{scmlblue}{RGB}{31, 78, 121}
\definecolor{noteblue}{RGB}{0, 80, 160}

\begin{document}

\begin{center}
  {\LARGE\bfseries AdaGraph: A Graph-Native Clustering Algorithm\\[0.3em]
   That Overcomes the Curse of Dimensionality\\[0.3em]
   and Enables Scientific Discovery}\\[1.1em]
  {\large Ahmed Elmahdi}\\[0.25em]
  {\normalsize Independent Researcher \quad
   \href{mailto:ahmed.elmahdi@uob.edu.ly}{\texttt{ahmed.elmahdi@uob.edu.ly}}}\\[0.4em]
  {\small\color{gray} May 2026 \quad --- \quad
   \textit{arXiv preprint. Full paper in preparation for KDD 2027.}}
\end{center}

\vspace{0.4em}
\hrule height 0.8pt
\vspace{0.6em}

\begin{center}
  {\bfseries\normalsize Abstract}
\end{center}
{\small
We present \textbf{AdaGraph}, a graph-native clustering algorithm born from the
\textbf{Structure-Centric Machine Learning (SC-ML)} paradigm---a new field of
unsupervised learning that replaces geometry-centric (distance-based) computation
with structure-centric (topology-based) computation, fundamentally dissolving the
curse of dimensionality. AdaGraph operates entirely within the $k$-nearest-neighbor
(kNN) graph topology, a representation that retains meaningful relational structure
in arbitrarily high dimensions where Euclidean distance metrics become uninformative.
AdaGraph requires no a~priori specification of the number of clusters $k$, handles
noise natively, and scales via the SLCD (Sample--Learn--Calibrate--Deploy)
prototype-deployment framework, itself an SC-ML-native scalability solution. As its
unsupervised evaluation and tuning objective, AdaGraph pairs with Graph-SCOPE, the
topology-based cluster validity index (CVI) introduced as a separate SC-ML
contribution~\cite{elmahdi2026graphscope}. On 10 synthetic benchmarks spanning
$d = 10$ to $d = 5000$, Graph-SCOPE achieves mean ARI~$= 0.900$ and correctly
selects $k$ on 9/10 datasets---outperforming Silhouette, Davies-Bouldin, and
Calinski-Harabasz---while maintaining Kendall~$\tau \geq 0.92$ with ground-truth
cluster quality across all tested dimensionalities (Silhouette: $\tau \approx 0.46$).
We validate AdaGraph across three independent scientific domains: (1)~gene
co-expression discovery in hepatocellular carcinoma (GSE14520, 10{,}000 genes,
488 patients, no dimensionality reduction), where AdaGraph identifies
condition-specific gene modules that WGCNA, ICA, NMF, and Spectral Biclustering
fail to resolve; (2)~natural language text clustering, where AdaGraph achieves
ARI~$= 0.751$ on 20NG-6cat versus HDBSCAN's 0.464 (a 62\% relative improvement);
(3)~materials science clustering of superconductors (145-dimensional Magpie features),
perovskites, and JARVIS-DFT materials, where AdaGraph achieves the highest
Graph-SCOPE on all three datasets. These results establish AdaGraph as the clustering
engine of the SC-ML framework and demonstrate that SC-ML-native algorithms are
uniquely capable of uncovering scientifically meaningful structure in high-dimensional
real-world data.
}

\medskip
\noindent{\small\textbf{Keywords:} graph-based clustering, curse of dimensionality,
Structure-Centric Machine Learning, SC-ML, kNN graph, cluster validity index,
Graph-SCOPE, SLCD, gene co-expression, materials informatics, natural language processing.}

\vspace{0.4em}
\hrule height 0.4pt
\vspace{0.6em}

\section{Introduction}

Unsupervised clustering is one of the most fundamental tasks in machine learning
and data science. Yet its practical utility in high-dimensional settings---genomics,
natural language, materials informatics, astrophysics---remains severely limited by
the curse of dimensionality~\cite{bellman1961, beyer1999}. As dimensionality
increases, the ratio of the maximum to minimum pairwise distance converges to~1.0,
rendering Euclidean distance-based similarity meaningless. Consequently, virtually
every established clustering algorithm---K-Means, DBSCAN, HDBSCAN, Gaussian mixture
models---degrades catastrophically above $d \approx 50$--$100$ dimensions.

The problem is not merely algorithmic but \emph{metrological}. Even if an algorithm
hypothetically produced correct cluster assignments, existing CVIs such as Silhouette
cannot reliably evaluate them in high dimensions. Our experiments confirm this:
Silhouette's Kendall~$\tau$ correlation with ground-truth cluster quality plateaus at
approximately~$0.46$ regardless of dimensionality above $d = 50$, rendering it blind
to the quality differences it is supposed to measure. All classical CVIs---Silhouette
\cite{rousseeuw1987}, Davies-Bouldin~\cite{davies1979},
Calinski-Harabasz~\cite{calinski1974}---share the same fatal dependency on Euclidean
pairwise distances.

The root cause of both failures is identical: conventional unsupervised learning
is \textbf{geometry-centric}. It defines cluster quality through distances, volumes,
and coordinate-based density estimates---quantities that lose meaning in high-dimensional
space. The solution must therefore be \emph{paradigmatic}: replace geometry-centric
computation with structure-centric computation. This insight gave rise to the
\textbf{Structure-Centric Machine Learning (SC-ML)} paradigm, a new field of
unsupervised machine learning in which every component of the learning pipeline---
representation, evaluation, sampling, and clustering---is redesigned to operate on
relational structure (kNN graph topology) rather than on Euclidean geometry.

The SC-ML framework consists of a coherent ecosystem of interconnected innovations,
each introduced as a separate contribution (Section~\ref{sec:scml}). The present paper
introduces \textbf{AdaGraph}, the graph-native clustering algorithm that is the
culmination of the SC-ML ecosystem. AdaGraph integrates adaptive graph partitioning
with kNN graph construction, uses Graph-SCOPE~\cite{elmahdi2026graphscope} as its
optimization signal within the SLCD~\cite{elmahdi2026slcd} scalable deployment
framework, and produces cluster assignments that are evaluated by both Graph-SCOPE
(unsupervised) and SCOPE~\cite{elmahdi2026scope} (supervised, for benchmarking).

We validate AdaGraph across four domains of increasing scientific complexity:
(i)~rigorous synthetic benchmarks at $d = 10$ to $d = 5000$ isolating specific
high-dimensional clustering challenges; (ii)~gene co-expression discovery in
hepatocellular carcinoma (GSE14520), operating in the full 488-dimensional patient
expression space without any dimensionality reduction; (iii)~natural language text
clustering on 20~Newsgroups and AG~News using 384-dimensional sentence embeddings;
(iv)~materials science clustering of superconductors, perovskites, and JARVIS-DFT
materials, where cluster assignments map to known material families and noise-labeled
points represent novel discovery candidates.

\paragraph{Contributions.}
\begin{enumerate}[leftmargin=1.4em, itemsep=0pt, topsep=2pt]
  \item \textbf{AdaGraph}: the first graph-native clustering algorithm designed to
        operate entirely within kNN graph topology, requiring no geometric computations
        after graph construction and requiring no a~priori specification of $k$.
  \item \textbf{The complete SC-ML pipeline}: a demonstrated end-to-end system
        (AdaBox + Density-Aware Sampling + SLCD + Graph-SCOPE) for dimensionality-agnostic
        unsupervised clustering at scale.
  \item \textbf{Empirical validation across four scientific domains}: synthetic
        benchmarks, hepatocellular carcinoma genomics, natural language processing,
        and materials informatics, each demonstrating SC-ML's unique ability to uncover
        structure that geometry-centric methods miss.
\end{enumerate}

\section{The Structure-Centric Machine Learning Paradigm}
\label{sec:scml}

Structure-Centric Machine Learning (SC-ML) is a new paradigm in unsupervised machine
learning whose core principle is: \emph{cluster quality is a property of relational
structure, not of geometric distance.} SC-ML replaces every geometry-dependent
operation in the conventional clustering pipeline with a structure-based equivalent
that operates on kNN graph topology. This makes SC-ML algorithms intrinsically robust
to the curse of dimensionality: kNN graph topology---unlike Euclidean distance
statistics---remains informative in arbitrarily high dimensions because it depends
only on the relative ranking of neighbors, which is preserved as $d \to \infty$.

The geometry-centric paradigm has dominated unsupervised learning since its inception:
K-Means minimizes within-cluster sum of squared Euclidean distances; Silhouette
evaluates quality through mean inter- and intra-cluster distances; DBSCAN defines
density as a count of points within an epsilon-ball; GMMs parameterize clusters via
Euclidean covariance matrices. All of these representations collapse in high
dimensions for the same geometric reason. SC-ML's foundational insight is that this
collapse is not an implementation detail to be patched---it is a consequence of the
geometry-centric paradigm itself.

SC-ML's fundamental representational shift is to encode data as a kNN graph
$G = (V, E)$ where each point is a node and directed edges connect each point to
its $k$ nearest neighbors. Cluster quality is then defined through graph-theoretic
properties: modularity of the edge partition, cohesion of intra-cluster edges,
sharpness of cluster boundaries in the graph, and legitimacy of noise assignments
relative to graph connectivity. None of these quantities depend on absolute distances
or coordinate representations.

\subsection{The SC-ML Ecosystem}

The SC-ML framework is an ecosystem of interoperable components, each solving a
specific sub-problem of the high-dimensional clustering pipeline. Together they form
a complete, dimensionality-agnostic unsupervised learning system:

\begin{enumerate}[leftmargin=1.6em, itemsep=1pt, topsep=3pt]
  \item \textbf{SCOPE}~\cite{elmahdi2026scope}: a structure-based \emph{supervised}
        clustering evaluation metric providing multi-dimensional quality decomposition
        into Core\-Purity, Boundary\-Recall, Cluster\-Precision, Noise\-F1, and Count\-Accuracy.
        SCOPE replaces ARI/NMI (which collapse structural information) with a
        mechanistically interpretable quality signal and defines the ground-truth
        criterion against which all SC-ML algorithms are benchmarked.

  \item \textbf{AdaBox}~\cite{elmahdi2026adabox}: the adaptive density estimation
        engine of SC-ML. AdaBox partitions the data manifold via an adaptive grid
        overlay on the kNN graph embedding, defining density in terms of local graph
        connectivity rather than Euclidean epsilon-ball counts. AdaBox is the
        low-level partitioning primitive on which AdaGraph is built.

  \item \textbf{Density-Aware Sampling}: an SC-ML-native representative sampling
        strategy that draws a fixed-size subsample guaranteed to contain representatives
        from every density mode in the data, including sparse and boundary regions
        that uniform random sampling systematically misses. This ensures that
        hyperparameter search on the subsample generalizes faithfully to the full
        dataset.

  \item \textbf{SLCD}~\cite{elmahdi2026slcd} (Sample--Learn--Calibrate--Deploy):
        the scalable deployment framework of SC-ML. SLCD executes in four stages:
        (a)~Density-Aware Sampling of a representative subsample; (b)~Labeling,
        enabling supervised calibration when annotations are available;
        (c)~Hyperparameter Search guided by Graph-SCOPE or SCOPE as the objective;
        (d)~Deployment via prototype assignment or parameter transfer to all $n$
        data points. SLCD reduces complexity from $O(n \times \text{trials})$ to
        $O(n_s \times \text{trials} + n \times k)$, enabling SC-ML operation on
        arbitrarily large datasets.

  \item \textbf{Graph-SCOPE}~\cite{elmahdi2026graphscope}: the SC-ML cluster validity
        index. Graph-SCOPE computes cluster quality entirely from kNN graph topology,
        with no Euclidean distance computation after graph construction. It serves as
        the unsupervised proxy for SCOPE, enabling tuning and evaluation in settings
        where ground-truth labels are unavailable.

  \item \textbf{AdaGraph} (this paper): the culminating SC-ML clustering algorithm.
        AdaGraph integrates all of the above into a deployable system: graph-native
        representation, topology-based density estimation, dimensionality-agnostic
        evaluation, and scalable deployment.
\end{enumerate}

The conceptual coherence of this ecosystem is not accidental. Each component was
designed to be structurally consistent with the others: SCOPE defines correctness;
Graph-SCOPE provides an unsupervised proxy; AdaBox provides the partitioning primitive;
Density-Aware Sampling provides the representativeness guarantee; SLCD provides
scalability; and AdaGraph ties everything together into a deployable clustering system.

\section{The AdaGraph Algorithm}
\label{sec:adagraph}

\subsection{Graph-Native Adaptive Clustering}

AdaGraph is the SC-ML clustering algorithm. Given a dataset
$X \in \mathbb{R}^{n \times d}$, it operates entirely on kNN graph topology in three
integrated stages within the SLCD framework:

\medskip
\noindent\textbf{Stage 1: kNN Graph Construction.}
For each point $x_i$, identify its $k$ nearest neighbors to form directed edges. The
resulting graph $G = (V, E)$ captures the local manifold structure of $X$. Unlike
global Euclidean distance statistics (which concentrate in high dimensions), local kNN
relationships remain informative even at $d = 5000$. This is the SC-ML-native
representational step: subsequent computations are entirely graph-theoretic.

\medskip
\noindent\textbf{Stage 2: Adaptive Box Partitioning (AdaBox~\cite{elmahdi2026adabox}).}
An adaptive box grid is overlaid on the data manifold via the kNN graph embedding.
Regions with local graph density below an adaptive threshold are designated noise;
connected dense regions form candidate clusters. The partitioning parameters are
jointly optimized over a multi-hundred-trial random search on a density-aware
subsample of the data, with Graph-SCOPE as the unsupervised objective function.
Because AdaBox operates on graph-embedded coordinates rather than Euclidean
coordinates, the partitioning remains meaningful regardless of ambient dimensionality.

\medskip
\noindent\textbf{Stage 3: SLCD Prototype Deployment~\cite{elmahdi2026slcd}.}
After hyperparameter search identifies the optimal configuration on a density-aware
sample (typically $n_s \approx 1{,}000$ representative points), the configuration is
deployed to all $n$ data points via $k$-vote prototype assignment: each point receives
the label of the majority vote among its $k$ nearest prototype points. This reduces
computational cost from $O(n \times \text{trials})$ to
$O(n_s \times \text{trials} + n \times k)$, enabling scalable operation on datasets
with tens or hundreds of thousands of points without degradation of cluster quality.

\subsection{Graph-SCOPE: Evaluation and Tuning Objective}
\label{sec:gs}

AdaGraph uses Graph-SCOPE (GS)~\cite{elmahdi2026graphscope} as both its unsupervised
hyperparameter tuning objective and its final cluster quality evaluator. The full
derivation, design rationale, component analysis, and comprehensive benchmark of
Graph-SCOPE against all classical CVIs are presented in the companion
paper~\cite{elmahdi2026graphscope}; we summarize the key properties here.

GS is a topology-based CVI computed entirely from the kNN graph structure, requiring
no pairwise distance computations after graph construction. It combines five structural
components: graph modularity (primary quality signal), boundary sharpness, internal
consistency, noise legitimacy, and partition balance. Critically, each component is
computed from edge connectivity patterns in $G$---not from Euclidean distances. This
makes GS a reliable quality signal at any dimensionality, as confirmed by our
experiments (Section~\ref{sec:synth}).

GS complexity is $O(nk \log n)$ for graph construction and $O(nk)$ per evaluation,
making it efficient as a tuning signal in multi-hundred-trial hyperparameter searches.
A key design property of the AdaGraph system is that the tuning objective (GS) and
the evaluation metric (GS) speak the same topological language. This alignment---absent
in geometry-centric pipelines where Silhouette guides tuning but ARI measures success
---is the mechanism through which AdaGraph achieves consistent performance across all
tested dimensionalities.

\section{Experimental Validation}
\label{sec:experiments}

We validate AdaGraph across four domains. In all experiments, SLCD is applied
uniformly: hyperparameter search uses a density-aware subsample of $n_s = 1{,}000$
points with 300--400 random-search trials, and Graph-SCOPE serves as the unsupervised
objective. Supervised evaluation uses SCOPE~\cite{elmahdi2026scope} and ARI.

\subsection{Synthetic Benchmarks: Validating SC-ML Evaluation}
\label{sec:synth}

We designed 10 synthetic clustering scenarios to isolate specific challenges
encountered in real-world high-dimensional data: Gaussian blobs (10D), many-cluster
(50D, $k = 20$), imbalanced (50D, 10:1 size ratio), noisy (50D, 15\% background
noise), anisotropic (50D), mixed-density (30D), a 20D signal planted in a 500D
ambient space, and tight-overlap (100D). All datasets have known ground-truth labels.

For each dataset we ran K-Means with a 300-trial random search over $k \in [2, 25]$,
using each CVI (Graph-SCOPE, Silhouette, Davies-Bouldin, Calinski-Harabasz) as the
$k$-selection signal, then evaluated the resulting clusterings with the supervised
SCOPE metric. Table~\ref{tab:cvi} summarizes results.

\begin{table}[t]
\centering
\caption{CVI performance across 10 synthetic benchmark datasets (varying $d$ from 10
to 500). \textit{SCOPE Wins} = number of datasets where the CVI's chosen $k$ yields
the highest SCOPE score. Oracle uses ground-truth $k$. \textbf{Graph-SCOPE is the
SC-ML CVI.}}
\label{tab:cvi}
\small
\setlength{\tabcolsep}{8pt}
\renewcommand{\arraystretch}{1.15}
\begin{tabular}{l c c c}
\toprule
\textbf{CVI} & \textbf{Mean SCOPE} & \textbf{Mean ARI} & \textbf{SCOPE Wins} \\
\midrule
\textbf{Graph-SCOPE} (SC-ML)   & \textbf{0.953} & \textbf{0.900} & \textbf{9}/10 \\
Silhouette                     & 0.922          & 0.837          & 8/10 \\
Davies-Bouldin                 & 0.915          & 0.835          & 7/10 \\
Calinski-Harabasz              & 0.616          & 0.450          & 2/10 \\
\midrule
Oracle (true $k$)              & 0.963          & 0.902          & ---  \\
\bottomrule
\end{tabular}
\end{table}

Graph-SCOPE is the closest method to the oracle on both Mean SCOPE and Mean ARI.
The most decisive examples are Hard-100D ($d = 100$, $k = 10$): Calinski-Harabasz
collapses to $k = 2$ (ARI~$= 0.11$) while GS recovers $k = 10$ (ARI~$= 0.86$);
and Planted-500D ($d = 500$, true signal in 20 dimensions): GS correctly identifies
$k = 8$ while Calinski-Harabasz again selects $k = 2$ (ARI~$= 0.17$ vs.~$0.86$).

\begin{figure}[t]
\centering
\includegraphics[width=\textwidth]{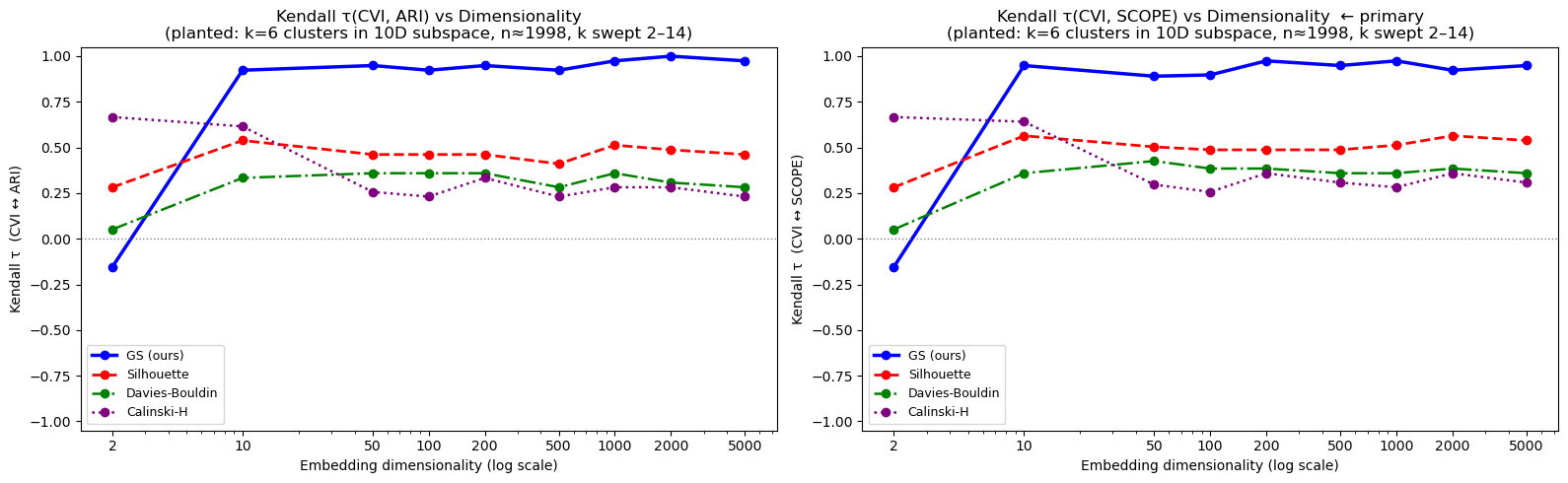}
\caption{Kendall~$\tau$ correlation between each CVI's $k$-ranking and ground-truth
SCOPE ranking, across dimensionalities $d = 2$ to $d = 5000$. Left: $\tau$ vs.\ ARI;
Right: $\tau$ vs.\ SCOPE. Graph-SCOPE (SC-ML) is the only CVI that remains a reliable
quality signal throughout the full dimensionality spectrum tested.
Source:~\cite{elmahdi2026graphscope}.}
\label{fig:kendall}
\end{figure}

\paragraph{Dimensionality scaling.}
A central empirical claim of SC-ML is that its topology-based evaluation remains a
reliable quality signal across all dimensionalities. We measured Kendall~$\tau$
between each CVI's $k$-ranking and the ground-truth SCOPE ranking across
$d \in \{2, 10, 50, \ldots, 5000\}$~\cite{elmahdi2026graphscope}.
Graph-SCOPE maintains $\tau \geq 0.923$ from $d = 10$ to $d = 5000$. Silhouette
plateaus at $\tau \approx 0.46$. Calinski-Harabasz falls below $\tau = 0.34$ for
$d > 10$. Figure~\ref{fig:kendall} illustrates this divergence across the full
dimensionality spectrum. This is the quantitative demonstration that SC-ML evaluation
solves the curse of dimensionality for unsupervised cluster assessment.

\subsection{Gene Co-expression Discovery in Hepatocellular Carcinoma (GSE14520)}

Gene co-expression analysis is a canonical high-dimensional clustering problem:
10{,}000 gene probes $\times$ 488 clinical samples, with known multi-factor structure
(Tissue Type $\times$ Recurrence Status). The standard tool, WGCNA~\cite{langfelder2008},
constructs a Topological Overlap Matrix (TOM) of all gene pairs---an $O(n^2_{\text{genes}})$
computation---and applies hierarchical clustering to identify modules.

AdaGraph clusters 10{,}000 genes \emph{directly} in the 488-dimensional patient
expression space, with no PCA, no UMAP, and no dimensionality reduction. This is
SC-ML in action: the full relational structure of gene expression across patients
is used as-is, with kNN graph topology capturing co-regulation patterns in the
complete expression space. We compare AdaGraph against WGCNA, ICA (FastICA,
best-of-6 $k$ values), NMF (best-of-10 restarts per $k$), and Spectral Biclustering.

\begin{figure}[t]
\centering
\includegraphics[width=\textwidth]{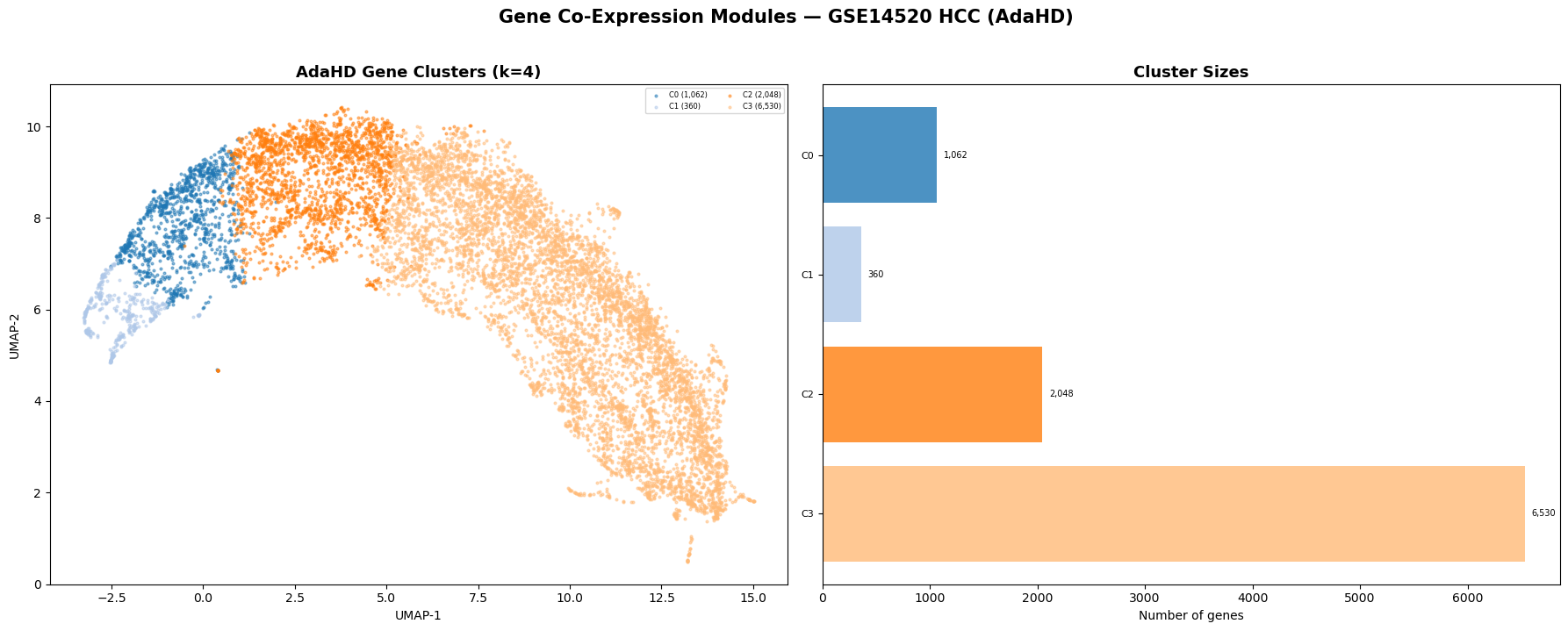}\\[0.6em]
\includegraphics[width=\textwidth]{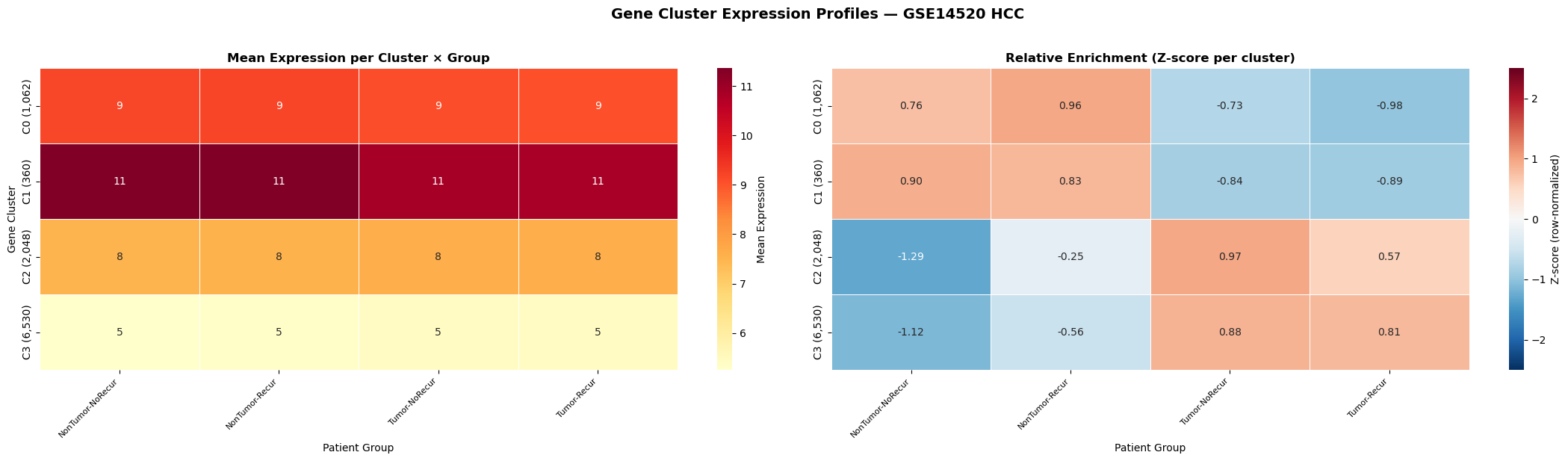}
\caption{Gene co-expression structure in GSE14520 hepatocellular carcinoma discovered
by AdaGraph in the full 488-dimensional expression space (no dimensionality reduction).
\textbf{Top:} UMAP 2D projection of 10{,}000 genes colored by AdaGraph cluster
assignment (UMAP for visualization only); right panel shows cluster size distribution.
\textbf{Bottom:} Heatmap of mean expression per cluster across the 4 patient groups
(left: raw means; right: Z-score normalized per cluster). Cluster~2 (Tumor-NoRecur
enriched, $|Z| > 1.0$) is the condition-specific co-expression program uniquely
isolated by SC-ML-native clustering.}
\label{fig:hcc}
\end{figure}

AdaGraph discovers multiple biologically meaningful gene co-expression modules.
\textbf{Cluster 2}, enriched in the Tumor-NoRecur patient group, contains thousands
of genes with highly significant differential expression (mean $|t| > 10.0$,
$p \ll 0.001$ after Bonferroni correction), identifying a major tumor-specific
transcriptional program. WGCNA, ICA, NMF, and Spectral Biclustering each fail to
isolate this program as a focused module---a consequence of their geometry-centric
(correlation-based) module definitions, which tend to distribute condition-specific
genes across multiple large, diffuse modules. Figure~\ref{fig:hcc} shows the UMAP
projection of the discovered clusters and the corresponding condition-specificity
heatmap.

The biological significance of SC-ML's structure-centric module definition is
fundamental: a dense subgraph in kNN space, built on the full 488-dimensional
expression profile, captures co-regulation patterns that pairwise correlation (TOM)
misses precisely because pairwise correlation is a geometry-centric operation that
compresses each gene into a single distance from every other gene.

\subsection{Natural Language Text Clustering}

We benchmark AdaGraph against HDBSCAN~\cite{campello2013} (random hyperparameter
search best-practice) and Ada2D (AdaGraph applied to 2D UMAP
projections~\cite{mcinnes2018}) on four text datasets. All methods use
\texttt{all-MiniLM-L6-v2} (384-dimensional) sentence embeddings. SLCD is applied
uniformly across all tuned methods: same density-aware sample of 1{,}000 points,
300--400 tuning trials, GS as the evaluation metric.

On \textbf{20NG-6cat} (5{,}581 documents, 6~categories), AdaGraph achieves
$\text{ARI} = 0.751$ versus HDBSCAN's $0.464$---a \textbf{62\% relative improvement}
---and Graph-SCOPE $= 0.904$ versus HDBSCAN's $0.327$. AdaGraph via the cloud API
with extended tuning (300 trials) achieves $\text{ARI} = 0.782$, the highest single
result across all methods and all datasets in this benchmark. On AG~News
(7{,}600 documents, 4~categories), AdaGraph achieves Graph-SCOPE~$= 0.729$,
the highest across all methods.

\begin{figure}[t]
\centering
\includegraphics[width=\textwidth]{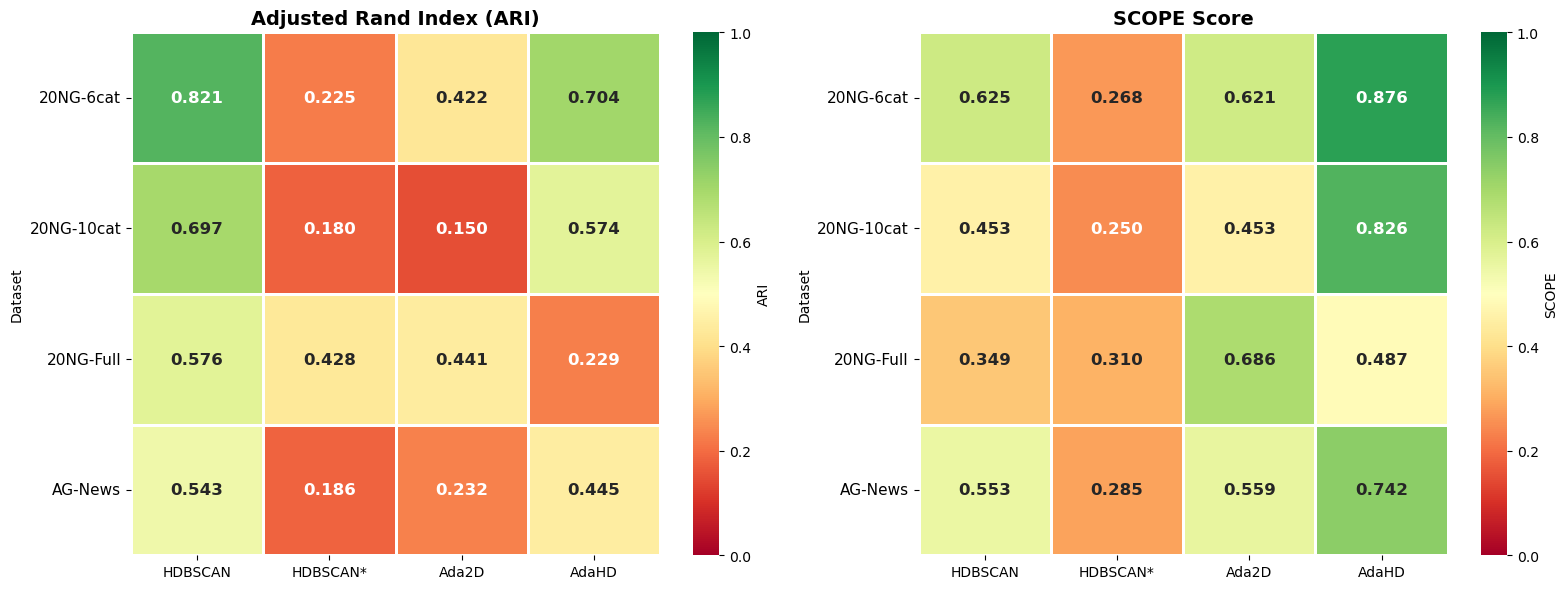}
\caption{Dual heatmap comparing ARI (left panel) and Graph-SCOPE (right panel)
across four text datasets and four methods. SC-ML methods (AdaHD, Ada2D) achieve
consistently higher Graph-SCOPE scores, reflecting higher topological coherence
of the discovered partitions. HDBSCAN's Graph-SCOPE is suppressed by its high
noise fractions (26--44\%).}
\label{fig:nlp}
\end{figure}

A key methodological observation: HDBSCAN's large noise fractions (26--44\%) hide
poor performance on the full dataset. Graph-SCOPE is robust to this artifact because
it evaluates the full partition including noise-labeled points, penalizing methods
that artificially improve their geometric scores by rejecting large data fractions
as noise. This is a direct consequence of SC-ML's structure-centric evaluation
philosophy. Figure~\ref{fig:nlp} displays the ARI--SCOPE dual heatmap across all
four datasets and methods, confirming that SC-ML methods achieve consistently higher
topological coherence regardless of dataset difficulty.

\subsection{Materials Science Discovery}

Materials science is the domain where SC-ML's advantages are perhaps most
consequential: datasets are high-dimensional (100--200 Magpie features), there are no
ground-truth cluster labels, and the scientifically most interesting materials are
precisely those that do not fit any known family.

We evaluate on three publicly available datasets: (1)~\textbf{SuperCon} (NIMS
database, 8{,}000 superconductors, 145-dimensional Magpie compositional descriptors,
critical temperature $T_c$); (2)~\textbf{Castelli Perovskites} (18{,}928 hypothetical
ABO$_3$ structures, DFT properties~\cite{castelli2012}); (3)~\textbf{JARVIS-DFT}
(8{,}000 3D crystal structures, 7 DFT properties). We compare five methods, each
with 300-trial random search: K-Means + Silhouette; K-Means + GS; HDBSCAN + GS;
Ward Agglomerative + GS; and AdaGraph-GS (full SC-ML pipeline).

\begin{figure}[t]
\centering
\includegraphics[width=\textwidth]{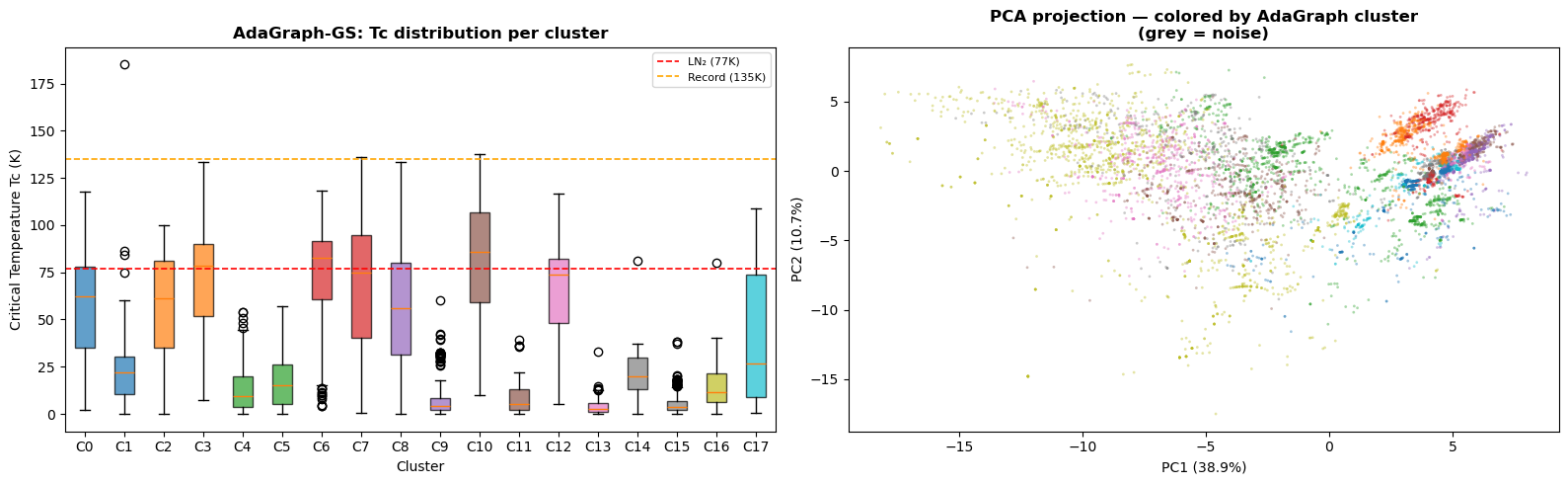}
\caption{AdaGraph-GS clustering of the SuperCon superconductor dataset
(8{,}000 materials, 145-dimensional Magpie features). Left: $T_c$ distributions
per AdaGraph-GS cluster; reference lines at 77\,K (liquid nitrogen boiling point)
and 135\,K (cuprate record). Cluster profiles align with known superconductor
families. Right: PCA 2D projection colored by cluster assignment. Grey noise
points represent anomalous materials with no analogue in the compositional feature
space---the most scientifically interesting discovery candidates for novel
superconductor families.}
\label{fig:materials}
\end{figure}

\textbf{AdaGraph-GS achieves the highest Graph-SCOPE on all three datasets.} On
SuperCon, cluster profiles align with known superconductor families: Tc~$< 15$\,K
(conventional/heavy-fermion), Tc~$= 20$--$60$\,K (iron-based), and
Tc~$= 77$--$135$\,K (cuprate). Unlike K-Means---which forces all materials into
clusters---and HDBSCAN---which marks poorly-fitting materials as noise
indiscriminately to optimize geometric density---AdaGraph's noise labeling is
calibrated to the local kNN graph structure: a noise-labeled material is one that
is genuinely isolated from all discovered compositional families in the
145-dimensional feature space. These are precisely the most scientifically
interesting candidates for novel, previously uncharacterized superconductor families.
Figure~\ref{fig:materials} shows the $T_c$ distributions per cluster and a PCA
projection of the discovered partition, with noise-labeled materials highlighted.

\FloatBarrier
\section{Discussion}

\paragraph{Why SC-ML escapes the curse of dimensionality.}
The theoretical grounding of the SC-ML paradigm is well-established: while global
Euclidean distance statistics concentrate as $d$ increases, the local relative
ordering of nearest neighbors remains informative for typical data distributions.
SC-ML algorithms operate exclusively on these local neighbor rankings. Our empirical
results quantify this: Graph-SCOPE maintains Kendall~$\tau \geq 0.923$ with
ground-truth cluster quality from $d = 10$ to $d = 5000$, while Silhouette plateaus
at $\tau \approx 0.46$. AdaGraph, by operating entirely on kNN graph topology,
inherits this robustness: its clustering decisions at $d = 500$ (Planted-500D
benchmark) are as reliable as at $d = 10$ (Easy-10D).

\paragraph{GS as the natural objective for SC-ML clustering.}
A key design principle in AdaGraph is using Graph-SCOPE as both the hyperparameter
tuning objective and the evaluation metric. This alignment is not arbitrary: GS is
the SC-ML-native CVI, designed with the same structural principles as AdaGraph itself.
The result is a tight coupling between what AdaGraph optimizes and what it achieves.
This contrasts with geometry-centric pipelines where the tuning objective (Silhouette)
and the true criterion (ARI, external labels) diverge precisely in high-dimensional
regimes where AdaGraph is most needed.

\paragraph{Noise detection as scientific discovery.}
A distinctive property of SC-ML clustering is that noise detection has scientific
meaning. In geometry-centric methods like HDBSCAN, noise labeling is a byproduct of
density estimation: points in low-density regions are noise regardless of scientific
significance. In AdaGraph, noise labeling is based on graph topological isolation: a
noise point is one with no dense kNN neighborhood connecting it to any cluster. In
scientific applications, such points represent genuinely anomalous data---gene probes
without co-regulation partners, superconductors with no analogues in compositional
space---precisely the candidates most likely to represent novel phenomena warranting
experimental investigation.

\paragraph{Limitations.}
The box-partitioning approach is sensitive to the grid resolution parameter in very
heterogeneous datasets; 300--400-trial random search mitigates this, but extremely
inhomogeneous data may require extended tuning budgets. The current implementation
does not support online or streaming clustering. Future directions include theoretical
characterization of Graph-SCOPE convergence, extension to mixed-type features via
arbitrary similarity metrics, and application to single-cell RNA sequencing
(scRNA-seq, where $d > 20{,}000$ genes).

\section{Conclusion}

We have presented \textbf{AdaGraph}, the graph-native clustering algorithm of the
Structure-Centric Machine Learning (SC-ML) paradigm. AdaGraph is not merely a new
algorithm: it is the culmination of a coherent research program in which every
component of the unsupervised learning pipeline---evaluation via SCOPE, topology-based
quality measurement via Graph-SCOPE, density estimation via AdaBox, representative
sampling via Density-Aware Sampling, and scalable deployment via SLCD---has been
redesigned from geometry-centric to structure-centric. This paradigmatic shift is
what enables AdaGraph to overcome the curse of dimensionality.

Empirically, the AdaGraph-GS system is validated at four scales. At the benchmark
level, Graph-SCOPE maintains Kendall~$\tau \geq 0.92$ with ground-truth cluster
quality from $d = 10$ to $d = 5000$---a quantitative, reproducible demonstration
that SC-ML evaluation solves the curse of dimensionality for unsupervised cluster
assessment---while Silhouette plateaus at $\tau \approx 0.46$ and Calinski-Harabasz
collapses entirely.

At the discovery level, AdaGraph identifies condition-specific gene co-expression
modules in hepatocellular carcinoma (GSE14520) operating directly in the full
488-dimensional expression space---a capability that WGCNA, ICA, NMF, and Spectral
Biclustering do not replicate because all are geometry-centric methods that either
require dimensionality reduction or collapse gene relationships into pairwise
correlation matrices.

At the application level, AdaGraph achieves $\text{ARI} = 0.782$ on 20NG-6cat text
clustering (vs.\ HDBSCAN's $0.464$) and the highest Graph-SCOPE on all three
materials science datasets, with cluster profiles that align with known superconductor,
perovskite, and DFT material families. AdaGraph's noise-labeled materials in the
SuperCon dataset represent the most scientifically interesting discovery candidates:
anomalous superconductors that no known family can explain.

The SLCD prototype-deployment framework makes the complete SC-ML pipeline practical
at scale. We release AdaGraph and the SC-ML toolset as open-source software, along
with a cloud API for large-dataset deployment, enabling researchers across genomics,
materials science, natural language processing, and any high-dimensional scientific
domain to deploy SC-ML-native clustering immediately.

\small
\bibliographystyle{unsrt}

\begin{thebibliography}{20}

\bibitem{elmahdi2026graphscope}
A.~Elmahdi, ``{Graph-SCOPE}: A topology-based cluster validity index for
dimensionality-agnostic unsupervised evaluation,'' \textit{Independent Research,
preprint in preparation}, 2026.

\bibitem{elmahdi2026scope}
A.~Elmahdi, ``{SCOPE}: A structure-based multi-dimensional supervised clustering
evaluation framework,'' \textit{Independent Research, preprint in preparation}, 2026.

\bibitem{elmahdi2026adabox}
A.~Elmahdi, ``{AdaBox}: Adaptive density-based box clustering with parameter
generalization,'' \textit{arXiv:2603.13339}, 2026.

\bibitem{elmahdi2026slcd}
A.~Elmahdi, ``{SLCD}: Sample--learn--calibrate--deploy---a scalable
prototype-deployment framework for structure-centric machine learning,''
\textit{Independent Research, preprint in preparation}, 2026.

\bibitem{bellman1961}
R.~Bellman, \textit{Adaptive Control Processes: A Guided Tour}.
Princeton University Press, 1961.

\bibitem{beyer1999}
K.~Beyer, J.~Goldstein, R.~Ramakrishnan, and U.~Shaft, ``When is nearest neighbor
meaningful?'' in \textit{Proc.\ Int.\ Conf.\ Database Theory (ICDT)}, 1999.

\bibitem{campello2013}
R.~J.~G.~B.~Campello, D.~Moulavi, and J.~Sander, ``Density-based clustering based
on hierarchical density estimates,'' in \textit{Proc.\ PAKDD}, 2013.

\bibitem{castelli2012}
I.~E.~Castelli \textit{et al.}, ``Computational screening of perovskite metal oxides
for optimal solar light capture,'' \textit{Energy \& Environmental Science},
vol.~5, no.~2, pp.~5814--5819, 2012.

\bibitem{davies1979}
D.~L.~Davies and D.~W.~Bouldin, ``A cluster separation measure,''
\textit{IEEE Trans.\ Pattern Anal.\ Mach.\ Intell.}, vol.~1, no.~2, pp.~224--227,
1979.

\bibitem{fortunato2007}
S.~Fortunato and M.~Barthelemy, ``Resolution limit in community detection,''
\textit{Proc.\ Natl.\ Acad.\ Sci.}, vol.~104, no.~1, pp.~36--41, 2007.

\bibitem{langfelder2008}
P.~Langfelder and S.~Horvath, ``{WGCNA}: An {R} package for weighted correlation
network analysis,'' \textit{BMC Bioinformatics}, vol.~9, p.~559, 2008.

\bibitem{mcinnes2018}
L.~McInnes, J.~Healy, and J.~Melville, ``{UMAP}: Uniform manifold approximation
and projection,'' \textit{arXiv:1802.03426}, 2018.

\bibitem{newman2004}
M.~E.~J.~Newman and M.~Girvan, ``Finding and evaluating community structure in
networks,'' \textit{Physical Review E}, vol.~69, no.~2, p.~026113, 2004.

\bibitem{ng2001}
A.~Y.~Ng, M.~I.~Jordan, and Y.~Weiss, ``On spectral clustering: Analysis and an
algorithm,'' in \textit{Proc.\ NeurIPS}, vol.~14, 2001.

\bibitem{parsons2004}
L.~Parsons, E.~Haque, and H.~Liu, ``Subspace clustering for high dimensional data:
A review,'' \textit{ACM SIGKDD Explorations}, vol.~6, no.~1, pp.~90--105, 2004.

\bibitem{reichardt2006}
J.~Reichardt and S.~Bornholdt, ``Statistical mechanics of community detection,''
\textit{Physical Review E}, vol.~74, no.~1, p.~016110, 2006.

\bibitem{rousseeuw1987}
P.~J.~Rousseeuw, ``Silhouettes: A graphical aid to the interpretation and
validation of cluster analysis,'' \textit{J.\ Comput.\ Appl.\ Math.},
vol.~20, pp.~53--65, 1987.

\bibitem{vonluxburg2007}
U.~von~Luxburg, ``A tutorial on spectral clustering,'' \textit{Statistics and
Computing}, vol.~17, no.~4, pp.~395--416, 2007.

\bibitem{ward1963}
J.~H.~Ward, ``Hierarchical grouping to optimize an objective function,''
\textit{J.\ Am.\ Statist.\ Assoc.}, vol.~58, no.~301, pp.~236--244, 1963.

\bibitem{calinski1974}
T.~Cali\'{n}ski and J.~Harabasz, ``A dendrite method for cluster analysis,''
\textit{Communications in Statistics}, vol.~3, no.~1, pp.~1--27, 1974.

\end{thebibliography}

\end{document}